\documentclass[runningheads]{utils/llncs}
\usepackage{graphicx}

\usepackage{tikz}
\usepackage{utils/comment}
\usepackage{amsmath,amssymb} 
\usepackage{color}
\usepackage{marvosym}
\usepackage{hyperref}
\usepackage{bm}
\usepackage{multirow}

\usepackage[accsupp]{axessibility}  

\begin{document}
\pagestyle{headings}
\mainmatter
\def\ECCVSubNumber{100}  

\title{Adversarial Feature Augmentation for Cross-domain Few-shot Classification} 

%
\author{Yanxu Hu\inst{1} \and
Andy J. Ma\inst{1,2,3}\textsuperscript{(\Letter)}\index{Ma, Andy J.}
}
\authorrunning{Y. Hu and A. J. Ma}
%
\institute{School of Computer Science and Engineering, Sun Yat-sen University, China \email{huyx69@mail2.sysu.edu.cn}, \email{majh8@mail.sysu.edu.cn} \and Guangdong Province Key Laboratory of Information Security Technology, China \and Key Laboratory of Machine Intelligence and Advanced Computing, Ministry of Education, China }

\maketitle

\begin{abstract}
Few-shot classification is a promising approach to solving the problem of classifying novel classes with only limited annotated data for training. 
Existing methods based on meta-learning predict novel-class labels for (target domain) testing tasks via meta knowledge learned from (source domain) training tasks of base classes.
However, most existing works may fail to generalize to novel classes due to the probably large domain discrepancy across domains.
To address this issue, we propose a novel adversarial feature augmentation (AFA) method to bridge the domain gap in few-shot learning.
The feature augmentation is designed to simulate distribution variations by maximizing the domain discrepancy.
During adversarial training, the domain discriminator is learned by distinguishing the augmented features (unseen domain) from the original ones (seen domain), while the domain discrepancy is minimized to obtain the optimal feature encoder.
The proposed method is a plug-and-play module that can be easily integrated into existing few-shot learning methods based on meta-learning.
Extensive experiments on nine datasets demonstrate the superiority of our method for cross-domain few-shot classification compared with the state of the art.
Code is available at \href{https://github.com/youthhoo/AFA_For_Few_shot_learning}{$https://github.com/youthhoo/AFA\_For\_Few\_shot\_learning$}.
\end{abstract}

\section{Introduction}
The development of deep convolutional neural networks (DCNNs) has achieved great success in image/video classification~\cite{HeZGLLLWW19,LiW0019,TanL19,WuHLLGD21}. 
The impressive performance improvement relies on the continuously upgrading computing devices and manual annotations of large-scale datasets.
To ease the heavy annotation burdens for training DCNNs, few-shot classification~\cite{lake2015human} has been proposed to recognize instances from novel classes with only limited labeled samples.
Among various recent methods to address the few-shot learning problem, the meta-learning approach~\cite{pmlr-v70-finn17a,FrikhaKKT21,LiuCLL0LH20,NIPS2017_cb8da676,SuiCMQLZ21,Sung_2018_CVPR,NIPS2016_90e13578,WuSLPZ20} have received a lot of attention due to its effectiveness. 
In general, meta-learning divides the training data into a series of tasks and learns an inductive distribution bias of these tasks to alleviate the negative impact of the imbalance between base and novel classes.  

\begin{figure}[t]
\begin{center}
\includegraphics[width=1\linewidth]{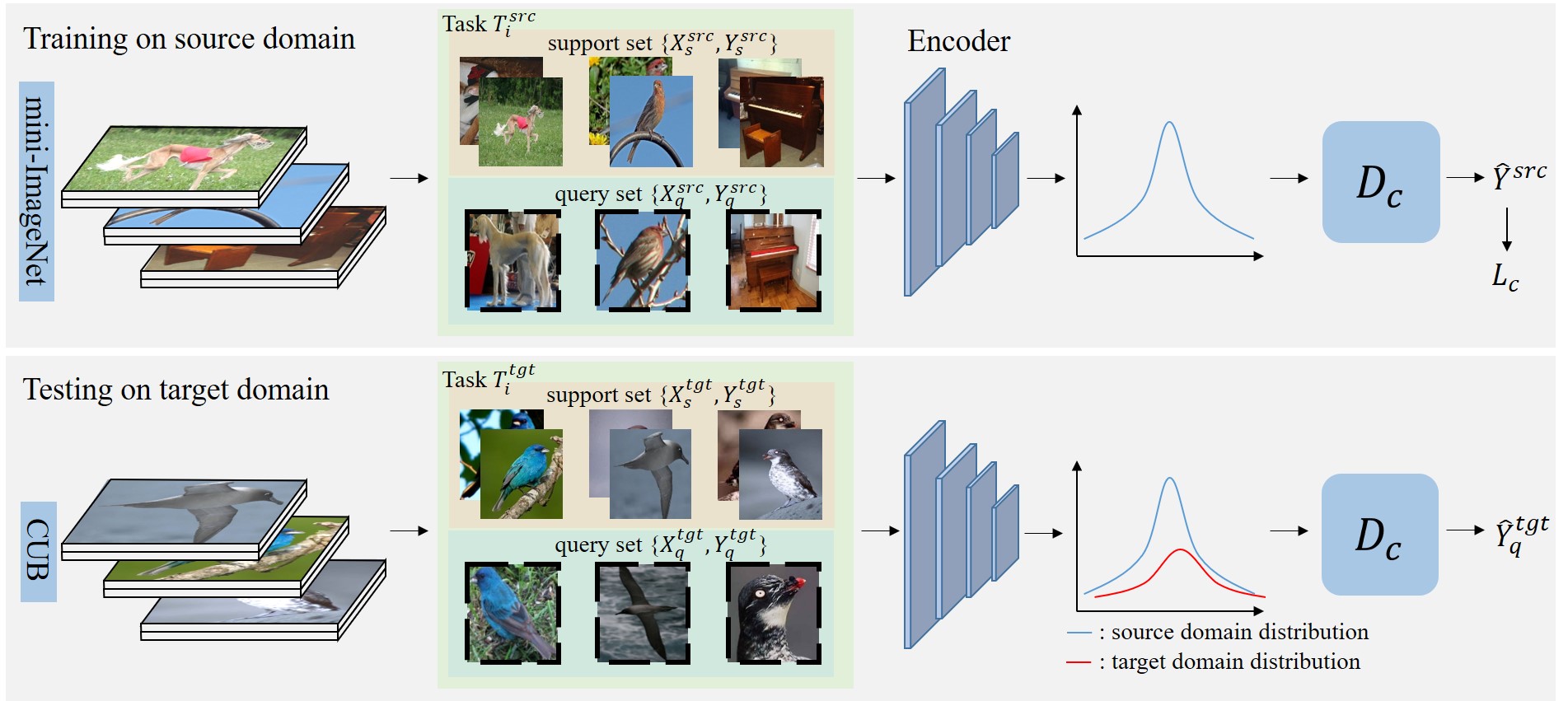}
\caption{{\bf Feature distribution misalignment problem in cross-domain few-shot classification.} 
In meta-learning methods, it consists of an feature encoder $E$ and a prediction head $D_c$.
There may be domain shift between the training (source domain) data of base classes and testing (target domain) data of novel classes. 
In this case, the distribution of the features extracted by the source domain encoder (\textcolor{blue}{blue}) differs from the target domain features (\textcolor{red}{red}).
Due to the distribution misalignment, {\textit{the meta-learned prediction head may not be able to correctly classify samples of novel classes from the target domain}}. 
Moreover, \textit{the feature distribution in the target domain can hardly be estimated due to the limited number of novel-class sample}.
In this paper, we propose a novel \textbf{{adversarial feature augmentation (AFA)}} method to learn domain-invariant features for cross-domain few-shot classification.
}
\label{figure0}
\end{center}
\end{figure}

Meta-learning is good at generalizing the base-class model to novel classes under the condition that the training distribution of base classes is almost equal to the testing one of novel classes.
Nevertheless, when the distributions of the training (source domain) and the testing (target domain) data differ from each other, the performance of the meta-learning model will degrade as justified by existing works~\cite{ChenLKWH19,GuoCKCSSRF20}.
Fig.~\ref{figure0} illustrates the domain shift problem in which the target dataset (e.g. CUB) is different from the source domain (e.g. mini-ImageNet).
In this scenario, the distribution of the target domain features extracted by the encoder $E$ may greatly deviate from the source domain distribution.

With the distribution misalignment, the class discriminator $D_c$ cannot make a correct decision for classifying novel-class data.
Domain adaptation (DA)~\cite{VolpiNSDMS18} can learn domain-invariant features by adversarial training~\cite{GaninUAGLLML17} to bridge the domain gap.
While DA assumes a lot of unlabelled samples are available in the target domain for training, the domain generalization (DG) approach~\cite{LiYZH19} can generalize from source domains to target domain without accessing the target data.
Differently, in few-shot learning, novel classes in the target domain do not overlap with base classes in the source domain and only very limited number of training samples are available for each class.
As a result, existing DA methods are not applicable for cross-domain few-shot classification.

To mitigate the domain shift under the few-shot setting, the adversarial task augmentation (ATA) method~\cite{WangD21} is proposed to search for the worst-case problem around the source task distribution.
While the task augmentation lacks of the capacity of simulating various feature distributions across domains, the feature-wise transformation (FT)~\cite{TsengLH020} is designed for feature augmentation using affine transforms.
With multiple source domains for training, the hyper-parameters in the FT are optimized to capture variations of the feature distributions.
When there is only single source domain, these hyper-parameters are empirically determined for training.
Though the FT achieves convincing performance improvement for both the base and novel classes, the empirical setting of hyper-parameters in the FT is sub-optimal.
Consequently, it cannot fully imitate the distribution mismatch under single source domain adaptation.

To overcome the limitations in existing works, we propose a novel adversarial feature augmentation (AFA) method for domain-invariant feature learning in cross-domain few-shot classification.
Different from the ATA, our method performs data augmentation in features (instead of tasks) to \textit{simulate the feature distribution mismatch across domains}.
Unlike the FT using multiple source domains to determine the optimal solution, the proposed AFA aligns the cross-domain feature distributions by \textit{adversarial learning based on single source domain}.

In our method, we design a feature augmentation module to transform the features extracted by the encoder $E$ according to sufficient statistics of normal distribution.
Considering the original and the augmented features as two different domains (seen and unseen respectively), the feature augmentation module is trained by maximizing the domain discrepancy across domains.
Moreover, the feature augmentation module is inserted into multiple layers of the encoder, such that the difference between the distributions of the seen and unseen domains are enlarged.
The distance between the gram matrices of multi-layer features from seen and unseen domains is used to measure the domain discrepancy.
During domain adversarial training, both the feature augmentation module and the domain discriminator is trained to distinguish the seen domain from the unseen one, while the encoder is learned by confusing the two different domains.


In summary, the contributions of this work are in three folds: 
\begin{enumerate}
\item We propose a model-agnostic feature augmentation module based on sufficient statistics of normal distribution.
The feature augmentation module can generate various feature distributions to better simulate the domain gap by maximizing the domain discrepancy under the cross-domain few-shot setting. 

\item We develop a novel adversarial feature augmentation (AFA) method for distribution alignment without accessing to the target domain data.
During adversarial training, the domain discriminator is learned by recognizing the augmented features (unseen domain) from the original ones (seen domain).
At the same time, the domain discrepancy is maximized to train the feature augmentation module, while it is minimized to obtain the optimal feature encoder.
In this way, the domain gap is reduced under the few-shot setting.

\item The proposed AFA is a plug-and-play module which can be easily integrated into existing few-shot learning methods based on meta-learning including matching net (MN)~\cite{NIPS2016_90e13578}, graph neural network (GNN)~\cite{SatorrasE18}, transductive propagation network (TPN)~\cite{LiuLPKYHY19}, and so on.
We experimentally evaluate the performance on the proposed method combined with the MN, GNN and TPN  under the cross-domain few-shot setting. 
Experimental results demonstrate that our method can improve the classification performance over the few-shot learning baselines and outperform the state-of-the-art cross-domain few-shot classification methods in most cases.
\end{enumerate}
\section{Related Work}

{\bf Few-shot classification.}
Few-shot classification~\cite{FinnAL17,FrikhaKKT21,GuoCKCSSRF20,LiuCLL0LH20,LiuLPKYHY19,WuSLPZ20} aims to recognize novel classes objects with few labeled training samples. MatchingNet~\cite{NIPS2016_90e13578} augments neural networks with external memories via LSTM module and maps a few labelled support samples and an unlabelled query samples to its label, while GNN~\cite{SatorrasE18} assimilates generic message-passing inference algorithms with their neural-network counterparts to interact the information between the labelled data and unlabelled data by graph. 
TPN~\cite{LiuLPKYHY19} learns a graph construction module that exploits the manifold structure in the data to propagate labels from labeled support images to unlabelled query instances, which can well alleviate the few-shot classification problem.
However, these meta-learning methods fail to generalize to target domains since the distribution of image features may vary largely due to the domain shift. Our work improves the generalization ability of the meta-learning model with the proposed adversarial feature augmentation (ATA) to better recognize target domain samples. 

{\noindent \bf Domain adaptation.}
Existing domain adaptation (DA) methods can be divided into three categories, i.e., discrepancy-based~\cite{LongC0J15,ZellingerGLNS17}, reconstruction-based approaches~\cite{DengSQZKPXL21,GhifaryKZBL16} and adversarial-based~\cite{GaninL15,GaninUAGLLML17,HsuYTHT0020,TzengHSD17}. For the discrepancy-based methods, DAN~\cite{LongC0J15} measures the distance between the distribution of source and target domain and the domain discrepancy is further reduced using an optimal multi-kernel selection method for mean embedding matching. The reconstruction-based method DRCN~\cite{GhifaryKZBL16} proposes a constructor to reconstruct target domain data, the more similar between the original data and constructed data, the more effective the feature learned by encoder are. While 
the adversarial-based method DANN~\cite{GaninUAGLLML17} learns domain-variance features by adversarial progress between encoder and domain discriminators. Nevertheless, these DA methods take the unlabelled data in the target domain as inputs for training, which is unavailable in the training stage under cross-domain few-shot classification setting.

{\noindent \bf Adversarial training.}
Adversarial training~\cite{GoodfellowSS14,MadryMSTV18,ShafahiNG0DSDTG19} is a powerful training module to improve the robustness of deep neural networks. 
To the end, Madry et al.~\cite{MadryMSTV18} develop projected gradient descent as a universal ``first-order adversary'' and use it to train model in adversarial way. 
Sinha et al.~\cite{SinhaND18} provide a training procedure that updates model parameters with worst-case perturbations of training data to perturb the data distribution, which has been referred by ATA~\cite{WangD21} to generate virtual `challenging' tasks to improve the robustness of models. In this work, we generate the ``bad-case perturbations'' in feature level via adversarial feature augmentation, which can simulate various feature distributions, to improve the generalization ability of various meta-learning methods.

{\noindent \bf Cross-Domain few-shot classification.}
Different from the few-shot domain adaptation works~\cite{SaitoKSDS19,YueZZ0DKS21}, the unlabelled data from target domain isn't used for training and the categories vary from training set to the testing set in cross-domain few-shot classification (CDFSC) problems. Compared to the few-shot classification, in the CDFSC, base classes are not share the same domain with novel classes. To improve the generalization of meta-learning methods, LRP~\cite{SunLSZCB20} develops a explanation-guided training strategy that emphasizes the features which are important for the predictions. While Wang et al.~\cite{WangD21} focus on elevating the robustness of various inductive bias of training data via enlarging the task distribution space. And Feature Transformation~\cite{TsengLH020} try to improve generalization to the target domain of metric-base meta-learning methods through modelling the various different distribution with feature-wise transformation layer.
Compared to the above methods, The CNAPs-based approaches~\cite{Requeima0BNT19,BateniBMW22,Bronskill0RNT20,BronskillMPHNT21,BateniGMWS20} developed from different perspectives, which is proposed based on Feature-wise Linear Modulation (FiLM) for efficient adaptation to new task at test time. 
Different from their approaches, we aim to simulate the various distributions in the feature-level with adversarial training and take it as feature augmentation to learn an encoder for extracting domain-invariant features.

\section{Proposed Method}
In this section, the preliminaries and the overall network architecture of our method are first introduced.
Then, the feature augmentation module and the adversarial training process are presented. 

\subsection{Preliminaries}
Following the few-shot classification setting~\cite{RaviL17}, the novel categories in the testing stage $C_{test}$ are different from base classes $C_{train}$ used in the training stage, i.e., $C_{train} \cap C_{test} = \emptyset$, then data for training and testing is divided into a series of tasks $T$. 
Each task contains a support set $T_s$ and a query set $T_q$. 
In a $n$-way $k$-shot support set $T_s$, the number of categories and labelled samples of each category is $n$ and $k$, respectively. 
The query set $T_q$ consists of the samples sharing the same classes as in $T_s$. 
During meta-learning, the training process on $T_s$ and the testing process on $T_q$ are called meta-training and meta-testing.
For each task, the goal is to correctly classify samples from $T_q$ by learning from $T_s$. 

With the domain shift problem in cross-domain few-shot classification, the training dataset (e.g. mini-ImageNet) is different from the testing data (e.g. CUB~\cite{welinder2010caltech} or Cars~\cite{Krause0DF13}).
In this work, we focus on adapting the meta model from single source domain to various target domains. 
In other words, only one source domain dataset is used for training while testing can be performed on different datasets. 
Notice that $T_s$ and $T_q$ of each task is from the same domain.
Since the labelled data from the target domain is very limited and the target novel classes are not overlapped with the source base classes, we propose to augment the features of each task by adversarial training to bridge the domain gap.

\begin{figure}[t]
\begin{center}
\includegraphics[width=1\linewidth]{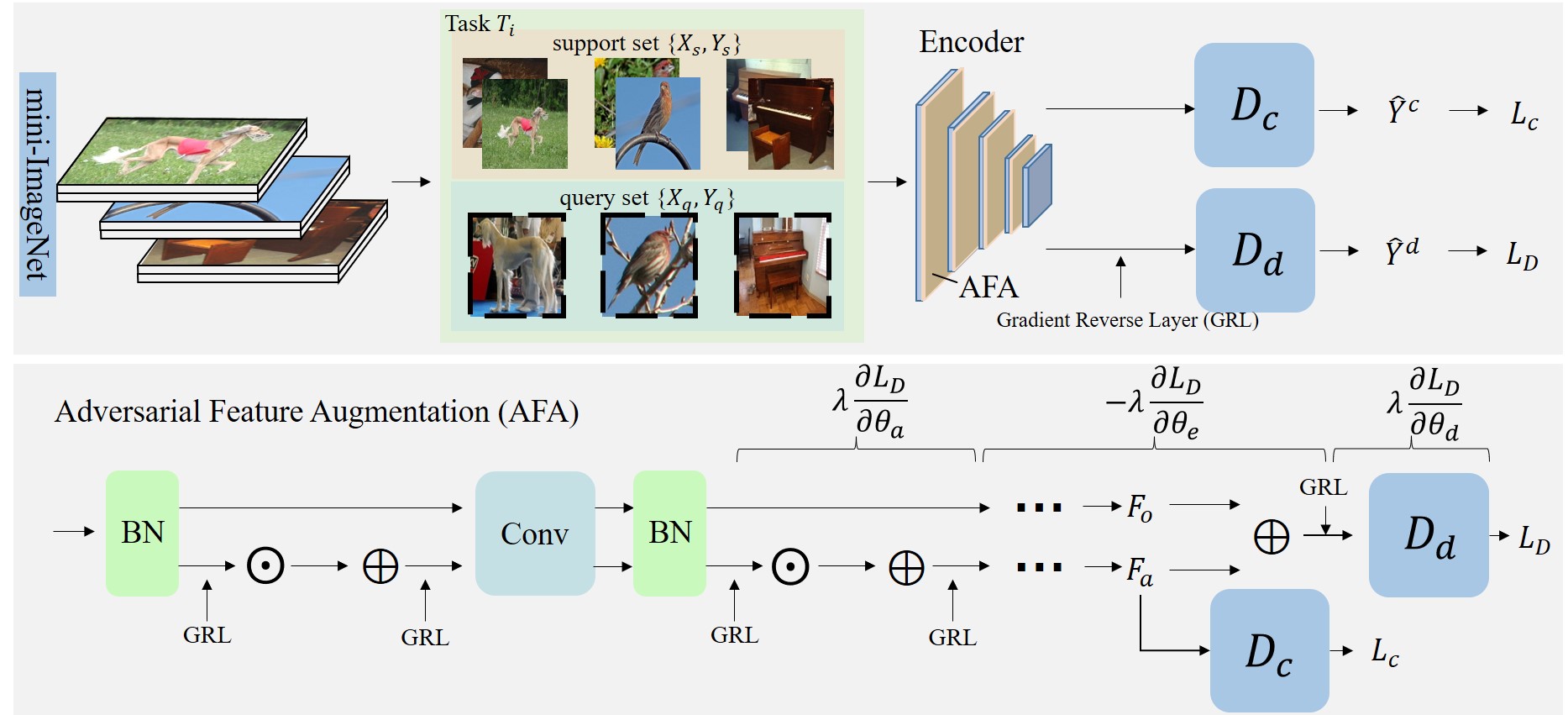}
\end{center}

   \caption{{\bf Top: Network Architecture.} 
   The network architecture of our method consists of a feature encoder $E$, a class discriminator and a domain discriminator $D_d$. 
   In the feature encoder $E$, a novel adversarial feature augmentation (AFA) module is embedded after each batch normalization layer.
   {\bf Bottom: Adversarial Feature Augmentation.} 
   The AFA module generates augmented features $F_a$ (unseen domain) to simulate distribution variations.
   By using adversarial training through inserting the gradient reverse layer (GRL) into the AFA module, the discrepancy between the distributions of $F_a$ and original features $F_o$ (seen domain) is maximized.
   At the same time, the domain discriminator $D_d$ is learned by distinguishing the seen domain from the unseen one, while the discrepancy $E$ is minimized to obtain the optimal feature encoder.
   Parameters of the AFA module, $D_d$ and $E$ are denoted as $\theta_a$, $\theta_e$ and $\theta_d$, respectively. 
   }
\label{figure1}
\end{figure}

\subsection{Network Architecture}
As shown in Fig.~\ref{figure1}, the network architecture of the proposed method contains a feature encoder $E$ and a class discriminator $D_c$ similar to meta-learning models.
Different from the traditional feature encoder, a novel adversarial feature augmentation (AFA) module is embedded after each batch normalization layer to simulate various feature distributions with details introduced in Section 3.3 \& 3.4.
With the augmented features (unseen domain), a domain discriminator $D_d$ is trained to distinguish the unseen domain from the seen one (original features).


The training procedures of our method follows the meta-learning approach to learn the inductive bias over the feature distribution from a series of tasks. 
By doing this, a class discriminator $D_c$ is learned and transferred to target tasks in the testing stage. 
For meta-training in each task, the base learner $\mathcal{B}$ outputs the optimal class discriminator $D_c$ based on the support set $T_s$ and the feature encoder $E$, i.e., $D_c = \mathcal{B}(E(T_s; \theta_e); \theta_c)$, where $\theta_c, \theta_e$ denote the learnable parameters of $D_c$ and $E$, respectively.
During meta-testing, the objective function is to minimize the classification loss of the query set $T_q$, i.e.,
\begin{equation}
\label{eq3}
\mathop{\min}\limits_{\theta_c, \theta_e} L_c = {L}_{c}({Y}_{q}^{c}, \hat{{Y}}_{q}^{c}), \hat{{Y}}_{q}^{c} = D_c(E(T_q; \theta_e); \theta_c) 
\end{equation}
where ${Y}_{q}^{c}$ and $\hat{{Y}}_{q}^{c}$ are the sets of ground-truth labels and predictions of the query images, respectively.
To mitigate the domain shift, we propose a novel AFA module integrated in the encoder $E$.
For each task, the output of $E$ in our method contains the original (seen domain) features $F_o\in \mathbb{R} ^ {N \times C}$ and augmented (unseen domain) features $F_a \in \mathbb{R} ^ {N \times C}$, where $N, C$ are the batch size and the number of channels, respectively. 
As shown in the bottom of Fig.~\ref{figure1}, $F_a$ representing the distribution varied from the source domain is used in the classification loss $L_c$.
When optimizing for the loss function $L_c$, the learnable parameters of the AFA module $\theta_a$ are fixed.
Details about how to learn the optimal $\theta_a$ and parameters $\theta_d$ of the domain discriminator $D_d$ are given in the following two subsections.


\subsection{Feature Augmentation}
To simulate various feature distributions, we design the feature augmentation function via disturbing the sufficient statistics of the original (seen domain) feature distribution.
Given a specified mean and variance, normal distribution best represents the current state of knowledge with maximum entropy.
As a results, we assume the feature maps in a training batch follows multivariate normal distribution and is independent and identically distributed.
Denote $f$ as any element in the feature map and ${f}_{1},\dots,{f}_{N}$ as the corresponding observations in a batch.
Since the marginal distribution of multivariate normal distribution is still normal, the probability density of a batch of $f$ in can be estimated by, 
\begin{equation}
\label{eq7}
p(f) = \prod_{i=1}^{N}\frac{1}{\sqrt{2\pi\sigma}}{\exp} \left({-\frac{({f}_{i}-\mu)^2}{2\sigma^2}}\right)
\end{equation}
where $\mu, \sigma$ are the mean and variance of $f$.
Then the probability density function can be decomposed into the part that is relevant to the overall distribution and is independent of the overall distribution,
By simplifying the product in right-hand side of Eq.~\eqref{eq7}, we have
\begin{equation}
\label{eq8}
p(f) 
=(2\pi\sigma)^{-\frac{N}{2}}{\exp} \left({-\frac{1}{2\sigma^2}\sum_{i}{({f}_{i}^2-2\mu{f}_{i}+\mu^2)}}\right)
\end{equation}

By Eq.~\eqref{eq8}, the probability density $p(f)$ can be decomposed into the form of a factor which does not depend on the distribution parameters $\mu, \sigma$ multiplied by the other factor depending on $\mu, \sigma$ and statistics $\sum_{i}{f}_{i}^2, \sum_{i}{f}_{i}$.
According to the Fisher–Neyman factorization theorem~\cite{fisher1922mathematical,splawa1990application}, $\sum_{i}{f}_{i}^2$ and $\sum_{i}{f}_{i}$ are sufficient statistics of normal distribution.
Moreover, the mean and variance are also sufficient statistics of normal distribution, because the statistics $\sum_{i}{f}_{i}^2$ and $\sum_{i}{f}_{i}$ can be calculated by them. 
The sufficient statistics of feature distribution include all the information of the distribution.
Thus, we propose to simulate various feature distributions by disturbing the mean and variance of the original features.

For this purpose, we insert a linear perturbation function with learnable parameters after each batch normalization layer.
Denote the original intermediate features from a certain batch normalization layer as ${m}^{o}\in\mathbb{R} ^ {C \times H \times W}$, where $H, W$ are the spatial resolutions of the feature map.
We initialize the scaling parameter $\gamma \in \mathbb{R}^{C}$ (for variance perturbation) and bias term $\beta\in \mathbb{R}^{C}$ (for mean disturbance) by normal distribution similar to~\cite{TsengLH020}. 
Then, the augmented feature ${m}^{a} \in\mathbb{R} ^ {C \times H \times W}$ is computed by, 
\begin{equation}
\label{eq9}
{m}^{a}_{c,h,w} = \gamma_{c} \times {m}^{o}_{c,h,w} + \beta_{c}
\end{equation}
The learnable parameters $\gamma, \beta$ are optimized by adversarial training which will be elaborated in the next subsection.

\subsection{Adversarial Feature Augmentation}
The augmentation module would not explore the distribution space that would enable the encoder to better handle tasks from a completely different domain, if $\gamma$ and $\beta$ is directly learned by solving the optimization problem in Eq.~\eqref{eq3}.
In this case, the ATA module fails to simulate the domain variations for domain-invariant feature learning.
In our method, we optimize the parameters in the AFA module, the domain discriminator $D_d$ and the feature encoder $E$ by adversarial training.
Let us consider the original features $F_o$ and the augmented features $F_a$ as the seen and unseen domain, respectively.
Denote the domain label of the input features as ${y}^{d}_{i}$.
If the input features are from the seen domain, then ${y}^{d}_{i} = 0$. Otherwise, ${y}^{d}_{i} = 1$. 
As in the DANN~\cite{GaninUAGLLML17}, the domain discriminator $D_d(\cdot): \mathbb{R} ^ {C} \rightarrow [0, 1]$ is defined as a logistic regressor, i.e.,
\begin{equation}
\label{eq4}
\hat{y}^{d}={D}_{d}(F;\boldsymbol{\mu}, b) = h({\boldsymbol{\mu}}^{T}F + b), \boldsymbol{\mu} \in \mathbb{R} ^ {C}, F = F_o \text{ or } F_a
\end{equation}
where $h(\cdot)$ is the sigmoid function, $\boldsymbol{\mu}, b$ are learnable parameters in $D_d$, and $\hat{Y}^{d}$ is the predicted labeled of the original features or the augmented features. 
Then, the loss function given by cross-entropy is,
\begin{equation}
\label{eq5}
{L}_{d}(\hat{Y}^{d},{Y}^{d}) = \frac{1}{2N}\sum_i[-{y}^{d}_{i}\log({\hat{y}^{d}_{i}}) - (1-{y}^{d}_{i})\log(1 - \hat{y}^{d}_{i})], 1 \leq i \leq 2N
\end{equation}
where ${Y}^{d}, \hat{Y}^{d}$ are the sets of all the ground-truth and predicted domain labels.
The domain discriminator can be trained by minimizing the loss function Eq.~\eqref{eq5} to distinguish between the seen and unseen domains.

Besides the domain similarity measured by the final output features from the encoder in Eq.~\eqref{eq5}, 
we also measure the domain discrepancy of the AFA module inserted after each batch normalization layer.
The gram matrices representing domain information of $m_o$ and $m_a$ are calculated as follows,
\begin{equation}
\label{eq10}
\hat{m} = Flatten(m), \hat{m} \in \mathbb{R}^{C \times S}, S = HW
\end{equation}
\begin{equation}
\label{eq11}
G(m) =  \hat{m} \times \hat{m}^{T}, G(m) \in \mathbb{R}^{C \times C}, m = m_o \text{ or } m_a
\end{equation}
Then, the domain discrepancy between the intermediate features $m_o$ and $m_a$ is determined by the distance between $G({m}_{o})$ and $G({m}_{a})$, i.e.,
\begin{equation}
\label{eq12}
{L}_{g} = \frac{1}{4{S}^{2}{C}^{2}}\sum_{i,j}\left(G_{i,j}(m_a) - G_{i,j}(m_o)\right)^2
\end{equation}
By maximizing the gram-matrix loss $L_g$, the AFA module is trained to ensure that the augmented intermediate features in each layer are different from the original ones to better mimic the target feature distribution.

For adversarial training, the domain similarity loss $L_d$ is maximized and the domain discrepancy loss $L_g$ is minimized to learn the feature encoder $E$ for distribution alignment.
In summary, the optimization problem for adversarial training is given as follows,
\begin{equation}
\label{eq13}
\max_{\theta_e} \min_{\theta_d, \theta_a} {L}_{D} = {L}_{d}-{L}_{g}
\end{equation}
The min-max optimization problem in Eq.~\eqref{eq13} can be solved as gradient reverse layers (GRLs) introudced in the DANN~\cite{GaninUAGLLML17}, which reverse the gradients during back propagation.
As shown in the bottom of Fig~\ref{figure1}, the gradients of the domain discriminator $D_d$, the encoder $E$ and the AFA module are updated by $\lambda{\partial L_D}/{\partial \theta_d}$, $-\lambda{\partial L_D}/{\partial \theta_e}$ and $\lambda{\partial L_D}/{\partial \theta_a}$, respectively, where $\lambda$ is a hyper-parameter set empirically as in the DANN. 


\textbf{Comparing to FT~\cite{TsengLH020}.} 
Both the feature-wise transformation (FT)~\cite{TsengLH020} and our method aim at transforming image features to simulate various feature distributions.
Our method takes full advantages of the original and augmented features to explicitly bridge the domain gap by adversarial training.
Thus, the distribution variations can be imitated by using only single source domain for training.
Nevertheless, FT relies on multiple source domains to learn the optimal feature transformation parameters.
Under the single source domain setting, the transformation parameters are set as constants, such that FT may suffer from the problem of performance drop as shown in our experiments.

\textbf{Comparing to ATA~\cite{WangD21}.} 
The adversarial task augmentation (ATA) method employs  adversarial training to search for the worst-case tasks around the source task distribution.
In this way, the space of the source task distribution could be enlarged, so that it may be closer to the task distribution in the target domain. 
Nevertheless, the perturbation on source tasks would degrade the performance on the unseen classes of source domain compared to other competitive models. 
Different from task augmentation, we propose feature augmentation with adversarial training via the gradient reverse layers to learn domain-invariant features without the problem of performance degradation.
Moreover, the ATA may not be able to fully utilize the available information in which only one of the generated tasks or the original tasks is used for training. 
In our method, both the original and augmented feature are used to train the domain discriminator.
At the same time, the proposed gram-matrix loss helps to generate unseen augmented features through maximizing the difference compared to the original features. 
In addition, ATA is more computational expensive to find the worst-case tasks via gradient ascents as shown in the complexity comparison in the supplementary.




\section{Experiment}
\subsection{Implementation}
In this section, we evaluate the proposed adversarial feature augmentation (AFA) module inserted into the Matching Network (MN)~\cite{NIPS2016_90e13578}, Graph Neural Network (GNN)~\cite{SatorrasE18} and Transductive Propagation Network (TPN)~\cite{LiuLPKYHY19}. 
We compare our method with the feature-wise transformation (FT)~\cite{TsengLH020}, explanation-guide training (LRP)~\cite{SunLSZCB20} and Adversarial Task augmentation(ATA)~\cite{WangD21}.


\subsection{Experimental Setting}
\label{data}

\textbf{Datasets.}
In this work, nine publicly available benchmarks are used for experiments, i.e., mini-ImageNet~\cite{NIPS2016_90e13578}, CUB~\cite{welinder2010caltech}, Cars~\cite{Krause0DF13}, Places~\cite{zhou2017places}, Plantae~\cite{HornASCSSAPB18}, CropDiseases, EuroSAT, ISIC and ChestX. 
Following the experimental setting of previous works~\cite{TsengLH020,WangD21}, we split these datasets into train/val/test sets, which is further divided into ${k}$-shot-${n}$-class support sets and the same ${n}$-class query sets.
We use the mini-ImageNet dataset as the source domain, and select the models with best accuracy on the validation set of mini-ImageNet for testing.

\noindent
{\bf Implementation Details.}
Our model can be integrated into existing meta-learning methods, e.g., MN~\cite{NIPS2016_90e13578}, GNN~\cite{SatorrasE18}, TPN~\cite{LiuLPKYHY19}.
In these methods, we use the ResNet-10~\cite{HeZRS16} with the proposed AFA module as the feature encoder.
The scaling term $\gamma \thicksim N(\textbf{1}, softplus(0.5))$ and bias term $\beta \thicksim N(\textbf{0}, softplus(0.3))$ are sampled from normal distribution for initialization.
To ensure fair comparison with the FT~\cite{TsengLH020}, LRP~\cite{SunLSZCB20}, ATA~\cite{WangD21} and the baseline methods. 
We follow the training protocol from \cite{ChenLKWH19}.
Empirically, the proposed model is trained with the learning rate 0.001 and 40,000 iterations. 
The performance measure is the average of the 2000 trials with randomly sampled batches. 
There are 16 query samples and 5-way 5-shot/1-shot support samples for each trial. 

\noindent
{\bf Pre-trained feature encoder.} Before the few-shot training stage, we apply an additional pre-training strategy as in FT~\cite{TsengLH020}, LRP~\cite{SunLSZCB20} and ATA~\cite{WangD21} for fair comparison.
The pre-trained feature encoder is minimized by the standard cross-entropy classification loss on the 64 training categories (the same as the training categories in few-shot training) in the mini-ImageNet dataset.


\begin{table}[t]
    \centering
    \scriptsize
    \setlength{\tabcolsep}{0.75pt}
    \begin{tabular}{lccccccccc}
    \hline
         \multirow{2}*{Method/shot}& \multicolumn{2}{c}{CUB} & \multicolumn{2}{c}{Cars}&\multicolumn{2}{c}{Places} & \multicolumn{2}{c}{Planae}  \\
         &1-shot &5-shot &1-shot &5-shot &1-shot &5-shot &1-shot &5-shot  \\ \hline
         MN~\cite{NIPS2016_90e13578}& $35.89_{\pm0.5}$&$51.37_{\pm0.8}$ &$30.77_{\pm0.5}$ &$38.99_{\pm0.6}$ &$49.86_{\pm0.8}$ &$63.16_{\pm0.8}$ &$32.70_{\pm0.6}$ &$46.53_{\pm0.7}$  \\
         w/ FT~\cite{TsengLH020} &$36.61_{\pm0.5}$&$55.23_{\pm0.8}$ &$29.82_{\pm0.4}$ &$41.24_{\pm0.7}$ &$51.07_{\pm0.7}$ &$64.55_{\pm0.8}$ &$34.48_{\pm0.5}$ &$41.69_{\pm0.6}$  \\
         w/ ATA~\cite{WangD21} & $39.65_{\pm0.4}$&$57.53_{\pm0.4}$ &$32.22_{\pm0.4}$ &$45.73_{\pm0.4}$ &$53.63_{\pm0.5}$ &$67.87_{\pm0.4}$ &$36.42_{\pm0.4}$ &$51.05_{\pm0.4}$ \\
         Ours & \bm{$41.02_{\pm0.4}$}&\bm{$59.46_{\pm0.4}$} &\bm{$33.52_{\pm0.4}$} &\bm{$46.13_{\pm0.4}$} &\bm{$54.66_{\pm0.5}$} &\bm{$68.87_{\pm0.4}$} &\bm{$37.60_{\pm0.4}$} &\bm{$52.43_{\pm0.4}$}  \\ \hline
         GNN~\cite{SatorrasE18}&$44.40_{\pm0.5}$&$62.87_{\pm0.5}$ &$31.72_{\pm0.4}$ &$43.70_{\pm0.4}$ &$52.42_{\pm0.5}$ &$70.91_{\pm0.5}$ &$33.60_{\pm0.4}$ &$48.51_{\pm0.4}$  \\
         w/ FT~\cite{TsengLH020}&$45.50_{\pm0.5}$&$64.97_{\pm0.5}$ &$32.25_{\pm0.4}$ &$46.19_{\pm0.4}$ &$53.44_{\pm0.5}$ &$70.70_{\pm0.5}$ &$32.56_{\pm0.4}$ &$49.66_{\pm0.4}$ \\
         w/ LRP~\cite{SunLSZCB20}& $43.89_{\pm0.5}$&$62.86_{\pm0.5}$ &$31.46_{\pm0.4}$ &$46.07_{\pm0.4}$ &$52.28_{\pm0.5}$ &$71.38_{\pm0.5}$ &$33.20_{\pm0.4}$ &$50.31_{\pm0.4}$ \\
         w/ ATA~\cite{WangD21}&$45.00_{\pm0.5}$&$66.22_{\pm0.5}$ &$33.61_{\pm0.4}$ &$49.14_{\pm0.4}$ &$53.57_{\pm0.5}$ &$75.48_{\pm0.4}$ &$34.42_{\pm0.4}$ &$52.69_{\pm0.4}$  \\
         Ours&\bm{$46.86_{\pm0.5}$}&\bm{$68.25_{\pm0.5}$} &\bm{$34.25_{\pm0.4}$} &\bm{$49.28_{\pm0.5}$} &\bm{$54.04_{\pm0.6}$} &\bm{$76.21_{\pm0.5}$} &\bm{$36.76_{\pm0.4}$} &\bm{$54.26_{\pm0.4}$}  \\ \hline
         TPN~\cite{LiuLPKYHY19}&$48.30_{\pm0.4}$&$63.52_{\pm0.4}$ &$32.42_{\pm0.4}$ &$44.54_{\pm0.4}$ &$56.17_{\pm0.5}$ &$71.39_{\pm0.4}$ &$37.40_{\pm0.4}$ &$50.96_{\pm0.4}$  \\
         w/ FT~\cite{TsengLH020}&$44.24_{\pm0.5}$&$58.18_{\pm0.5}$ &$26.50_{\pm0.3}$ &$34.03_{\pm0.4}$ &$52.45_{\pm0.5}$ &$66.75_{\pm0.5}$ &$32.46_{\pm0.4}$ &$43.20_{\pm0.5}$  \\
         w/ ATA~\cite{WangD21}&$50.26_{\pm0.5}$&$65.31_{\pm0.4}$ &$34.18_{\pm0.4}$ &$46.95_{\pm0.4}$ &$57.03_{\pm0.5}$ &$72.12_{\pm0.4}$ &$39.83_{\pm0.4}$ &$55.08_{\pm0.4}$  \\
         Ours&\bm{$50.85_{\pm0.4}$}&\bm{$65.86_{\pm0.4}$} &\bm{$38.43_{\pm0.4}$} &\bm{$47.89_{\pm0.4}$} &\bm{$60.29_{\pm0.5}$} &\bm{$72.81_{\pm0.4}$} &\bm{$40.27_{\pm0.4}$} &\bm{$55.67_{\pm0.4}$}  \\ \hline
         & \multicolumn{2}{c}{CropDiseases} & \multicolumn{2}{c}{EuroSAT}&\multicolumn{2}{c}{ISIC} & \multicolumn{2}{c}{ChestX}  \\
         &1-shot &5-shot &1-shot &5-shot &1-shot &5-shot &1-shot &5-shot   \\ \hline
         MN~\cite{NIPS2016_90e13578}& $57.57_{\pm0.5}$&$73.26_{\pm0.5}$ &$54.19_{\pm0.5}$ &$67.50_{\pm0.5}$ &$29.62_{\pm0.3}$ &$32.98_{\pm0.3}$ &$22.30_{\pm0.2}$ &$22.85_{\pm0.2}$  \\
         w/ FT~\cite{TsengLH020} &$54.21_{\pm0.5}$&$70.56_{\pm0.5}$ &$55.62_{\pm0.4}$ &$63.33_{\pm0.5}$ &$30.64_{\pm0.3}$ &$35.73_{\pm0.3}$ &$21.50_{\pm0.2}$ &$22.88_{\pm0.2}$  \\
         w/ ATA~\cite{WangD21}& $55.57_{\pm0.5}$&$79.28_{\pm0.4}$ &$56.44_{\pm0.5}$ &$68.83_{\pm0.4}$ &$31.48_{\pm0.3}$ &\bm{$40.53_{\pm0.3}$} &$21.52_{\pm0.2}$ &\bm{$23.19_{\pm0.2}$}  \\
         Ours& \bm{$60.71_{\pm0.5}$}&\bm{$80.07_{\pm0.4}$} &\bm{$61.28_{\pm0.5}$} &\bm{$69.63_{\pm0.5}$} &\bm{$32.32_{\pm0.3}$} &$39.88_{\pm0.3}$ &\bm{$22.11_{\pm0.2}$} &$23.18_{\pm0.2}$  \\ \hline
         GNN~\cite{SatorrasE18}& $59.19_{\pm0.5}$&$83.12_{\pm0.4}$ &$54.61_{\pm0.5}$ &$78.69_{\pm0.4}$ &$30.14_{\pm0.3}$ &$42.54_{\pm0.4}$ &$21.94_{\pm0.2}$ &$23.87_{\pm0.2}$  \\
         w/ FT~\cite{TsengLH020}& $60.74_{\pm0.5}$&$87.07_{\pm0.4}$ &$55.53_{\pm0.5}$ &$78.02_{\pm0.4}$ &$30.22_{\pm0.3}$ &$40.87_{\pm0.4}$ &$22.00_{\pm0.2}$ &$24.28_{\pm0.2}$  \\
         w/ LRP~\cite{SunLSZCB20}& $59.23_{\pm0.5}$&$86.15_{\pm0.4}$ &$54.99_{\pm0.5}$ &$77.14_{\pm0.4}$ &$30.94_{\pm0.3}$ &$44.14_{\pm0.4}$ &$22.11_{\pm0.2}$ &$24.53_{\pm0.3}$  \\
         w/ ATA~\cite{WangD21}& $67.45_{\pm0.5}$&\bm{$90.59_{\pm0.3}$} &$61.35_{\pm0.5}$ &$83.75_{\pm0.4}$ &\bm{$33.21_{\pm0.4}$} &$44.91_{\pm0.4}$ &$22.10_{\pm0.2}$ &$24.32_{\pm0.4}$  \\
         Ours& \bm{$67.61_{\pm0.5}$}&$88.06_{\pm0.3}$ &\bm{$63.12_{\pm0.5}$} &\bm{$85.58_{\pm0.4}$} &\bm{$33.21_{\pm0.3}$} &\bm{$46.01_{\pm0.4}$} &\bm{$22.92_{\pm0.2}$} &\bm{$25.02_{\pm0.2}$}  \\ \hline
         TPN~\cite{LiuLPKYHY19}& $68.39_{\pm0.6}$&$81.91_{\pm0.5}$ &$63.90_{\pm0.5}$ &$77.22_{\pm0.4}$ &$35.08_{\pm0.4}$ &$45.66_{\pm0.3}$ &$21.05_{\pm0.2}$ &$22.17_{\pm0.2}$  \\
         w/ FT~\cite{TsengLH020}& $56.06_{\pm0.7}$&$70.06_{\pm0.7}$ &$52.68_{\pm0.6}$ &$65.69_{\pm0.5}$ &$29.62_{\pm0.3}$ &$36.96_{\pm0.4}$ &$20.46_{\pm0.1}$ &$21.22_{\pm0.1}$  \\
         w/ ATA~\cite{WangD21}&\bm{$77.82_{\pm0.5}$}&\bm{$88.15_{\pm0.5}$} &$65.94_{\pm0.5}$ &$79.47_{\pm0.3}$ &\bm{$34.70_{\pm0.4}$} &$45.83_{\pm0.3}$ &$21.67_{\pm0.2}$ &\bm{$23.60_{\pm0.2}$}  \\
         Ours& $72.44_{\pm0.6}$&$85.69_{\pm0.4}$ &\bm{$66.17_{\pm0.4}$} &\bm{$80.12_{\pm0.4}$} &$34.25_{\pm0.4}$ &\bm{$46.29_{\pm0.3}$} &\bm{$21.69_{\pm0.1}$} &$23.47_{\pm0.2}$  \\ \hline
    \end{tabular}
    \caption{Few-shot classification accuracy (\%) of 5-way 5-shot/1-shot setting trained on the mini-ImageNet dataset, and tested on various datasets from target domains. 
    The best results in different settings are in \textbf{Bold}. 
    }
    \label{tab:base}
\end{table}

\subsection{Results on Benchmarks}

We train each model using the mini-ImageNet as the source domain and evaluate the model on the other eight target domains, i.e., CUB, Cars, Places, Plantae, CropDiseases, EuroSAT, ISIC and ChestX.
In our method, the AFA module is inserted after each batch normalization layer of the feature encoder during the training stage. 
All the results are shown in Table 1. We have following observations from these results:
\textbf{\emph{i}.}
Our method outperforms the state of the art for almost all the datasets and different-shot settings in different meta-learning methods. 
For 1-shot classification, our method improves the baselines by ${3.45\%}$ averagely over the eight datasets in different models. 
In 5-shot setting, the average improvement is ${4.25\%}$ compared to the baselines.
\textbf{\emph{ii}.}
Compared to the competitive ATA~\cite{WangD21}, our method integrated with the proposed AFA achieves an average improvement of about ${1\%}$.

\subsection{Ablation Experiments}
\textbf{Effect of the domain discriminator.} 
As mentioned in Section 3.1, we apply the domain discriminator to maximize the discrepancy between the augmented features and original features. 
In this experiment, we perform ablation experiments of the domain discriminator through training the AFA via the classification loss function $L_c$ but without using the domain discriminator. 
The classification accuracy on various datasets are reported in the second line of Table~\ref{tab:domain}. 
Based on the results, we have the following observations:
\textbf{\emph{i}.}
When using the AFA without the domain discriminator, the performance degrades. 
This indicates that it can improve the performance on various datasets under the settings with different number of shots by training with the domain discriminator,  (e.g. an average $2.48\%$ improvement on the 1-shot setting).
\textbf{\emph{ii}.}
Compared to the baseline (MN), the AFA without the domain discriminator also help to generalize to various domains in most cases. 
These results demonstrate that the adversarial training can alleviate the antagonistic action between the class discriminator and the feature augmentation module.
\textbf{\emph{iii}.}
Although training with the AFA module with the classification loss can improve the performance in most datasets, it leads to a decline in the CropDiseases dataset comparing with the baseline.

\begin{table}[t]
    \centering
    \scriptsize
    \setlength{\tabcolsep}{0.5pt}
    \begin{tabular}{lccccccccc}
         \hline
         \multirow{2}*{Method/shot}& \multicolumn{2}{c}{CUB} & \multicolumn{2}{c}{Cars}&\multicolumn{2}{c}{Places} & \multicolumn{2}{c}{Planae}  \\
         &1-shot &5-shot &1-shot &5-shot &1-shot &5-shot &1-shot &5-shot  \\ \hline
         MN~\cite{NIPS2016_90e13578}& $35.89_{\pm0.5}$&$51.37_{\pm0.8}$ &$30.77_{\pm0.5}$ &$38.99_{\pm0.6}$ &$49.86_{\pm0.8}$ &$63.16_{\pm0.8}$ &$32.70_{\pm0.6}$ &$46.53_{\pm0.7}$  \\
         w/o $D_d$& $38.83_{\pm0.4}$&$58.06_{\pm0.4}$ &$32.35_{\pm0.4}$ &$45.92_{\pm0.4}$ &$51.46_{\pm0.5}$ &$65.45_{\pm0.4}$ &$36.80_{\pm0.4}$ &$49.00_{\pm0.4}$  \\
         w/o $L_g$& $40.49_{\pm0.4}$&$56.44_{\pm0.4}$ &$31.08_{\pm0.3}$ &$44.78_{\pm0.4}$ &$51.98_{\pm0.5}$ &$66.60_{\pm0.4}$ &$35.03_{\pm0.4}$ &$50.56_{\pm0.4}$  \\
         Non-linear &$34.42_{\pm0.4}$&$50.17_{\pm0.4}$ &$28.77_{\pm0.3}$ &$42.04_{\pm0.4}$ &$49.92_{\pm0.4}$ &$59.00_{\pm0.4}$ &$34.27_{\pm0.4}$ &$50.90_{\pm0.4}$  \\
         Ours & \bm{$41.02_{\pm0.4}$}&\bm{$59.46_{\pm0.4}$} &\bm{$33.52_{\pm0.4}$} &\bm{$46.13_{\pm0.4}$} &\bm{$54.66_{\pm0.5}$} &\bm{$68.87_{\pm0.4}$} &\bm{$37.60_{\pm0.4}$} &\bm{$52.43_{\pm0.4}$}  \\ \hline
         & \multicolumn{2}{c}{CropDiseases} & \multicolumn{2}{c}{EuroSAT}&\multicolumn{2}{c}{ISIC} & \multicolumn{2}{c}{ChestX}  \\
         &1-shot &5-shot &1-shot &5-shot &1-shot &5-shot &1-shot &5-shot   \\ \hline
         MN~\cite{NIPS2016_90e13578}& $57.57_{\pm0.5}$&$73.26_{\pm0.5}$ &$54.19_{\pm0.5}$ &$67.50_{\pm0.5}$ &$29.62_{\pm0.3}$ &$32.98_{\pm0.3}$ &$22.30_{\pm0.2}$ &$22.85_{\pm0.2}$  \\
         w/o $D_d$ & $56.87_{\pm0.5}$&$71.02_{\pm0.5}$ &$54.78_{\pm0.5}$ &$67.66_{\pm0.4}$ &$30.19_{\pm0.3}$ &$38.83_{\pm0.3}$ &$21.70_{\pm0.2}$ &$22.86_{\pm0.2}$  \\
         w/o $L_g$ & $55.43_{\pm0.5}$&$74.62_{\pm0.5}$ &$57.41_{\pm0.5}$ &$65.50_{\pm0.4}$ &$30.78_{\pm0.3}$ &$36.87_{\pm0.3}$ &$21.23_{\pm0.2}$ &$22.96_{\pm0.2}$  \\
         Non-linear &$56.01_{\pm0.5}$&$79.95_{\pm0.4}$ &$56.90_{\pm0.5}$ &\bm{$72.15_{\pm0.4}$} &$29.19_{\pm0.3}$ &\bm{$40.20_{\pm0.3}$} &$20.95_{\pm0.2}$ &\bm{$23.58_{\pm0.2}$} \\
         Ours& \bm{$60.71_{\pm0.5}$}&\bm{$80.07_{\pm0.4}$} &\bm{$61.28_{\pm0.5}$} &$69.63 _{\pm0.5}$ &\bm{$32.32_{\pm0.3}$} &$39.88_{\pm0.3}$ &\bm{$22.11_{\pm0.2}$} &$23.18_{\pm0.2}$  \\ \hline
    \end{tabular}
    \caption{Accuracy (\%) of ablation experiments under 1-shot/5-shot 5-way few-shot classification on the target domains datasets.
    \textbf{w/o $D_d$} is the experiment without the domain discriminator. 
    \textbf{w/o $L_g$} is the ablation experiment of the gram-matric loss calculated by Eq.~\eqref{eq10}. \textbf{Non-linear} indicate that the linear transformation of adversarial feature augmentation is replaced by convolution layer as the non-linear transformation. 
    Here we use the matching network (MN) as the baseline for experiments.}
    \label{tab:domain}
\end{table}
\noindent
\textbf{Effect of the gram-matrix loss.} 
The gram-matrix loss function is to measure the difference between augmented and original features in each AFA module. 
The accuracy on various datasets are reported on the third line of Table~\ref{tab:domain}. Compared with the last line results, we can find that the gram-matrix loss brings about $2.57\%$ improvement. 
It leads to the best results for novel classes by combining the domain discriminator and the gram-matrix loss. 
The main reason behind is that these two modules contribute to a complementary improvement on global and local discrepancy between the augmented and original features.

\noindent
\textbf{How about non-linear transformation?}
In Section 3.2, we introduce linear perturbation in the AFA module to mimic various feature distributions via disturbing the sufficient statistics of original feature distribution. 
Here, we replace the linear transformation with the non-linear transformation (convolution) layers to generate unseen feature distribution.
The classification accuracy are report in the forth line of Table~\ref{tab:domain}. 
As we can see, the non-linear transformation cannot bring obvious improvement or even performs worse. 
It verify the theoretical justification of our method based on sufficient statistic such that the disturbance to the mean and variance is better for generalizing to target domain.

\subsection{Results of base classes and novel classes}
\begin{figure}[t]
\begin{center}
\includegraphics[width=1\linewidth]{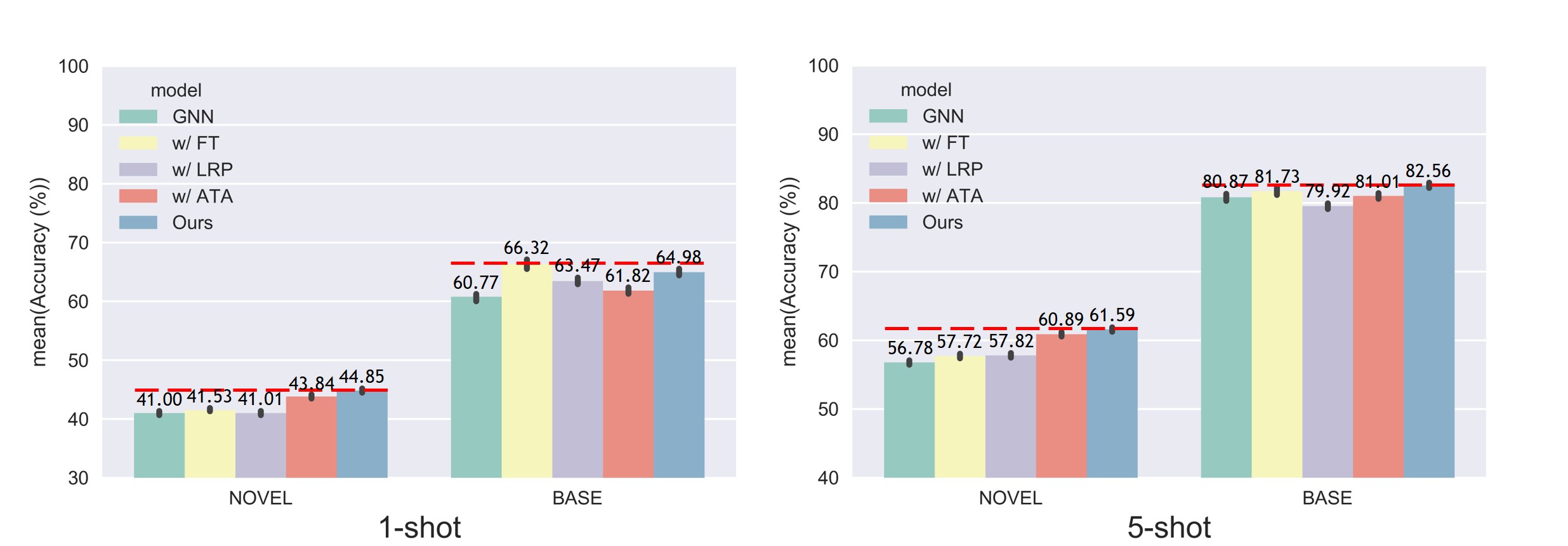}
\caption{Accuracy (\%) of baseline (GNN), FT, LRP, ATA and our model for 1/5-shot cross-domain classification on both novel classes and base classes. 
   }
\label{figure2}
\end{center}
\end{figure}

Since meta-learning methods may show inconsistent results for \textit{base} and \textit{novel} classes, we report results of both novel and base classes accuracy ($\%$) for comparison in Fig.~\ref{figure2}.
Here the performance of the novel classes is the average accuracy of the eight datasets, i.e., CUB, Cars, Places, Plantae, CropDiseases, EuroSAT, ISIC and ChestX.
The base classes are the rest categories of the mini-ImageNet dataset different from the categories used for training. 
Our proposed modules with GNN perform better than the baseline (GNN)  on both the base and novel classes, which indicates that the AFA module does not sacrifice the base classes performance to make do with cross-domain few-shot learning.
The red dashed line of the base classes on 1-shot setting show that the Graph Convolution Network (GNN) with feature-wise transformation~\cite{TsengLH020} has the slight improvement over our model on the base classes, but the performance on the novel classes degrades significantly. 
Moreover, compared to the competitive methods ATA~\cite{WangD21}, our method remarkably improve the performance on the base classes.
Our method suppresses all the related works by the performance on the novel classes and also achieves competitive results on base classes.
All these results demonstrate that our method can give the best balance between base and novel classes and classify the samples of novel classes well.

\begin{table}[t]
    \centering
    \scriptsize
    \setlength{\tabcolsep}{0.5pt}
    \begin{tabular}{lccccccccc}
         \hline
         \multirow{2}*{Method/shot}& \multicolumn{2}{c}{CUB} & \multicolumn{2}{c}{Cars}&\multicolumn{2}{c}{Places} & \multicolumn{2}{c}{Planae}  \\
         &1-shot &5-shot &1-shot &5-shot &1-shot &5-shot &1-shot &5-shot  \\ \hline
         Fine-tuning& $43.53_{\pm0.4}$&$63.76_{\pm0.4}$ &$35.12_{\pm0.4}$ &$51.21_{\pm0.4}$ &$50.57_{\pm0.4}$ &$70.68_{\pm0.4}$ &$38.77_{\pm0.4}$ &$56.45_{\pm0.4}$  \\
         MN+$Ours^*$ & $43.62_{\pm0.4}$&$68.73_{\pm0.4}$ &$36.83_{\pm0.4}$ &$52.53_{\pm0.4}$ &$52.82_{\pm0.5}$ &$71.56_{\pm0.4}$ &$38.56_{\pm0.4}$ &$56.50_{\pm0.4}$  \\
         GNN+$Ours^*$ & $47.40_{\pm0.5}$&\bm{$70.33_{\pm0.5}$} &$36.50_{\pm0.4}$ &\bm{$55.75_{\pm0.5}$} &$55.34_{\pm0.6}$ &\bm{$76.92_{\pm0.4}$} &$39.97_{\pm0.4}$ &\bm{$59.58_{\pm0.5}$}  \\
         TPN+$Ours^*$ & \bm{$48.05_{\pm0.5}$}&$67.78_{\pm0.4}$ &\bm{$38.45_{\pm0.4}$} &$54.89_{\pm0.4}$ &\bm{$57.27_{\pm0.5}$} &$73.06_{\pm0.4}$ &\bm{$40.85_{\pm0.4}$} &$59.04_{\pm0.4}$  \\ \hline
         & \multicolumn{2}{c}{CropDiseases} & \multicolumn{2}{c}{EuroSAT}&\multicolumn{2}{c}{ISIC} & \multicolumn{2}{c}{ChestX}  \\
         &1-shot &5-shot &1-shot &5-shot &1-shot &5-shot &1-shot &5-shot   \\ \hline
         Fine-tuning& $73.43_{\pm0.5}$&$89.84_{\pm0.3}$ &$66.17_{\pm0.5}$ &$81.59_{\pm0.3}$ &$34.60_{\pm0.3}$ &$49.51_{\pm0.3}$ &$22.13_{\pm0.2}$ &$25.37_{\pm0.2}$  \\
         MN+$Ours^*$ & $74.67_{\pm0.4}$&$90.53_{\pm0.3}$ &$66.48_{\pm0.5}$ &$82.00_{\pm0.3}$ &$34.58_{\pm0.3}$ &$48.46_{\pm0.3}$ &$22.29_{\pm0.2}$ &\bm{$25.80_{\pm0.3}$} \\
         GNN+$Ours^*$ &$74.80_{\pm0.5}$&\bm{$95.66_{\pm0.2}$} &$69.64_{\pm0.6}$ &\bm{$89.56_{\pm0.4}$} &\bm{$35.33_{\pm0.4}$} &\bm{$50.44_{\pm0.4}$} &$22.25_{\pm0.2}$ &$24.96_{\pm0.2}$  \\
         TPN+$Ours^*$ & \bm{$81.89_{\pm0.5}$}&$93.67_{\pm0.2}$ &\bm{$70.37_{\pm0.5}$} &$86.68_{\pm0.2}$ &$34.88_{\pm0.4}$ &$50.17_{\pm0.3}$ &\bm{$22.65_{\pm0.2}$} &$24.79_{\pm0.2}$  \\ \hline
    \end{tabular}
    \caption{Accuracy (\%) of fine-tuning with the augmented support dataset from the target domain and our model for 1/5-shot 5-way classification on the target domains. 
    $*$ means the method fine-tuned with target tasks generated through data augmentation. \textbf{Bold} indicates the best results.}
    \label{tab:finetune}
\end{table}
\subsection{Comparison with Fine-tuning}
As mentioned by Guo et al.~\cite{GuoCKCSSRF20}, when coming across the domain shift, traditional pre-training and fine-tuning methods perform better than meta-learning methods in few-shot setting. 
This experiment is to verify that the superiority of the meta-learning methods with our module over the traditional pre-training and fine-tuning under the cross-domain few-shot setting. 
For a fair comparison, we follow the way of Wang et al.~\cite{WangD21}, i.e, using data augmentation for fine-tuning in target tasks. 
Given an target task $T$ formed by the $k$-shot $n$-way samples as support set and $n \times 15$ pseudo samples as query set. 
The pseudo samples of query set are generated by the support samples using the data augmentation method from~\cite{abs09218}. 
For pre-training and fine-tuning, we first pre-train the model with the source tasks composed of the mini-ImageNet dataset. 
Then, the trained feature encoder is used for initialization and a fully connected layer is used as the discriminator to fulfill the unseen tasks mentioned above for fine-tuning. 
We use the SGD optimizer with learning rate 0.01 the same as~\cite{GuoCKCSSRF20}. 
For the meta-learning methods with our proposed module, we initialize the parameters of model with the meta-learning on the source tasks and then used the same support and query samples of the target task as above. 
We apply the Adam optimizer with the learning rate 0.001. 
Both fine-tuning and meta-learning methods are fine-tuned for 50 epoch under the 5-shot/1-shot 5-way setting. 
Since the data used for training are consistent in all models, it is a fair comparison. 
As shown in the Table 3, our method consistently outperforms the traditional pre-training and fine-tuning.



\section{Conclusions}

In this paper, we present a novel method namely Adversarial Feature Augmentation (AFA) which can generate augmented features to simulate domain variations and improve the generalization ability of meta-learning models. 
Based on sufficient statistics of normal distribution, the feature augmentation module is designed by perturbation on feature mean and variance.
By adversarial training, the ATA module is learned by maximizing the domain discrepancy with the domain discriminator, while the feature encoder is optimized by confusing the seen and unseen domains. 
Experimental results on nine datasets show that the proposed AFA improves the performance of meta-learning baselines and outperforms existing works for cross-domain few-shot classification in most cases. \\

\noindent\textbf{Acknowledgments.} This work was supported partially by NSFC (No.61906218), Guangdong Basic and Applied Basic Research Foundation (No.2020A1515011497), and Science and Technology Program of Guangzhou (No.202002030371).

\clearpage
%
%
\bibliographystyle{utils/splncs04}
\bibliography{utils/egbib}
\end{document}